%% file: root.tex
\documentclass[letterpaper, 10 pt, conference]{ieeeconf}  
\usepackage{amsmath} 
\usepackage{amsfonts}

\usepackage{graphicx}
\usepackage{hyperref}       
\usepackage{cleveref}

\usepackage{booktabs}
\usepackage{multirow}
\usepackage[table]{xcolor}
\usepackage{threeparttable}

\input{notation}

\IEEEoverridecommandlockouts                              

\usepackage{siunitx}

\usepackage{xcolor}
\definecolor{Adap}{RGB}{230,89,214} 
\definecolor{Tac}{RGB}{89,195,230}

\title{\LARGE \bf
Adaptive Visuo-Tactile Fusion with Predictive Force Attention for Dexterous Manipulation
}

\author{Jinzhou Li$^{*}$, Tianhao Wu$^{*}$, Jiyao Zhang$^{\dagger}$, Zeyuan Chen$^{\dagger}$, \\Haotian Jin, Mingdong Wu, Yujun Shen, Yaodong Yang, Hao Dong
\thanks{Jinzhou Li, Tianhao Wu, Jiyao Zhang, Zeyuan Chen, Haotian Jin, Mingdong Wu and Hao Dong are with the Center on Frontiers of Computing Studies, School of Computer Science, Peking University, also with PKU-Agibot Lab, School of Computer Science, Peking University, and also with National Key Laboratory for Multimedia Information Processing, School of Computer Science, Peking University. Yunjun Shen is with Ant Group. Yaodong Yang is with Institute for Artificial Intelligence, Peking University. }%
\thanks{* and $\dagger$ indicates equal contribution.}
\thanks{Corresponding to hao.dong@pku.edu.cn.}%
}

\def\figmk{Fig.~}
\def\tablemk{Tab.~}

\begin{document}

\maketitle
\thispagestyle{empty}
\pagestyle{empty}

\begin{abstract}
Effectively utilizing multi-sensory data is important for robots to generalize across diverse tasks. However, the heterogeneous nature of these modalities makes fusion challenging. Existing methods propose strategies to obtain comprehensively fused features but often ignore the fact that each modality requires different levels of attention at different manipulation stages. To address this, we propose a force-guided attention fusion module that adaptively adjusts the weights of visual and tactile features without human labeling. We also introduce a self-supervised future force prediction auxiliary task to reinforce the tactile modality, improve data imbalance, and encourage proper adjustment. Our method achieves an average success rate of 93\% across three fine-grained, contact-rich tasks in real-world experiments. Further analysis shows that our policy appropriately adjusts attention to each modality at different manipulation stages. The videos can be viewed at \url{https://adaptac-dex.github.io/}.

\end{abstract}

\section{INTRODUCTION}
\begin{figure}[t!]
    \centering
    \includegraphics[width=1.0\linewidth]{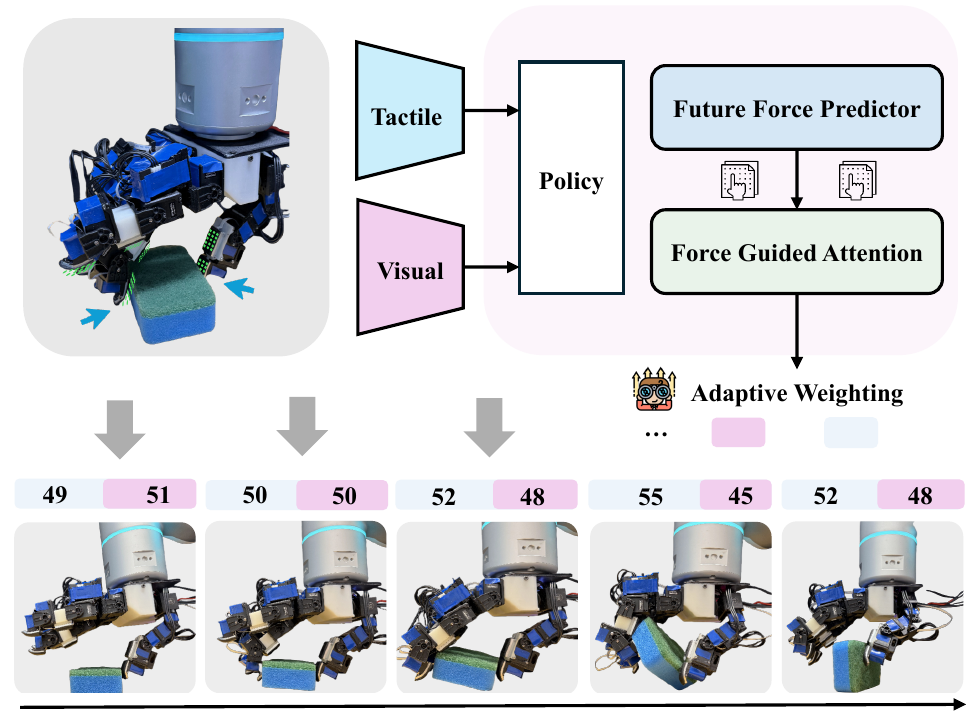}
    \caption{\textbf{Adaptive Visuo-Tactile Fusion with Predictive Force Attention.} Our policy leverages visual and tactile information to predict future force and combines it with the observed force to adaptively adjust the attention of different modalities at different stages of dexterous manipulation.}
    \label{fig:system}
    \vspace{-5mm}
\end{figure}

Humans rely on multiple senses, particularly vision and touch, to effectively perceive and interact within the physical world. Consider the task of flipping a dish sponge in \figmk\ref{fig:system}: we naturally use vision to quickly locate the object, and rely on touch to precisely adjust finger placement and apply suitable force during contact. Similarly, for robots to effectively perform such tasks, it is essential to understand \textit{when}, \textit{where}, and \textit{how} contact occurs, and to integrate this understanding with visual sensory information.
Recent studies have focused on integrating visual and tactile sensors into robotic hands and exploring visuo-tactile policies for dexterous manipulation~\cite{guzey2024see, lin2024learning, wu2024canonical, guzey2023dexterity}.

However, vision and touch exhibit fundamentally different characteristics. Vision is more global and provides contextual information, while tactile sensing is more local and offers precise feedback on physical contact. Effectively integrating these distinct sensory streams into a coherent understanding remains a significant challenge. Therefore, we ask the question: how can we design multisensory robotic policy frameworks that effectively bridge the heterogeneity between vision and touch, leveraging their complementary strengths?

Prior works have explored data-level fusion~\cite{yin2023rotating,hu2023dexterous,huang20243dvitac} that directly combines raw or preprocessed inputs from different sensors—for example, combining point clouds and tactile data within a unified 3D representation~\cite{huang20243dvitac,yin2024learning}. This straightforward strategy typically involves simple concatenation, helping to preserve fine-grained details from each modality. However, such data often exhibits heterogeneous characteristics, including differences in spatial resolution, sampling rates, and noise distributions. Direct concatenation can therefore introduce challenges with respect to aligning and integrating sensor data across these dimensions.

In addition, researchers have also explored feature-level fusion \cite{li2019connecting,dave2024multimodal,lin2024learning,liu2024masked}.  These works primarily encourage the network to fully leverage both unique and shared information from each modality to form a more comprehensive representation of the current state. However, they overlook that different modalities require varying levels of attention at different manipulation stages~\cite{badde2020modality}, and in some cases, additional modalities can even be a distraction~\cite{guzey2024see,he2024foar}. Predicting contact probability to modulate sensory features is one strategy to address such distractions. FoAR~\cite{he2024foar}, for instance, implements this concept by using its contact predictions to adjust force features from end-effector force/torque in real-time, helping to avoid noise interference. However, it relies on a predefined threshold to label contact and assumes that vision features always dominate, which limits its scalability.

In this work, we propose Adaptive Visuo-Tactile Fusion with Predictive Force Attention (\textbf{AdapTac}), a method for adaptively integrating visual and tactile modalities in dexterous manipulation. Unlike prior approaches that rely on fixed assumptions (e.g., vision dominance) or manually defined thresholds, AdapTac leverages contact-induced force that naturally reflects contact states and interaction dynamics. Specifically, 
1) AdapTac introduces a force-guided attention module, where force signals serve as queries, and visual/tactile features as keys and values. By computing attention weights between the force and different modalities, our policy adaptively adjusts the attention of visual and tactile, eliminating the need for task-specific labels and improving generalization. 
2) AdapTac also incorporates a self-supervised auxiliary loss, which uses a diffusion force head to predict future force signals during training. This auxiliary future force prediction reinforces the tactile modality and alleviates data imbalance. The predicted future forces, combined with observed force information, provide temporal context of contact to effectively guide visuo-tactile fusion, ensuring context-aware modality weighting and efficient attention adjustment.

In summary, our contributions are as follows:
(i) We propose a force-guided cross-attention fusion module that adaptively adjusts visual and tactile feature weighting using force signals, enabling flexible fusion without task-specific human labeling.
(ii) We introduce a self-supervised auxiliary task for future force prediction, and leverage predicted and observed forces to guide visuo-tactile fusion, improving data balance and efficiency.
(iii) We demonstrate the effectiveness and robustness of our approach through extensive real-world experiments and comprehensive ablation studies on visuo-tactile dexterous manipulation tasks.

\section{RELATED WORK}

\subsection{3D Imitation Learning}
Imitation Learning (IL) allows robots to learn policies by imitating expert demonstrations~\cite{NIPS1996_68d13cf2, JMLR:v17:15-522, duan2017one}. Recent advancements include the Diffusion Policy~\cite{chi2023diffusionpolicy}, which uses diffusion models~\cite{ho2020denoising} for diverse actions, and the Action Chunk Transformer (ACT)~\cite{zhao2023learning}, which predicts structured action chunks for long-term behavior. Additionally, the shift from RGB images~\cite{chi2023diffusionpolicy, pari2021surprising, zhao2023learning} to 3D geometric forms~\cite{huang20243dvitac, ze20243d, 10801678, chisari2024learning, wang2024dexcap} captures richer spatial information, improving the model's generalization. Early works with multi-view voxelized point clouds~\cite{shridhar2022perceiveractor, gervet2023act3d} laid the foundation for spatial understanding. More recently, the 3D Diffusion Policy~\cite{ze20243d} has further highlighted the potential of point clouds for action policies. However, such encoders like DP3~\cite{ze20243d}, PointNet++~\cite{qi2017pointnet++}, and Transformer variants models~\cite{jaegle2021perceiver} still struggle with noise and sparsity, limiting their real-world performance. RISE~\cite{10801678} improves robustness with sparse convolutional network~\cite{choy20194d}. Despite these advances, most IL methods rely heavily on visual data, overlooking the critical role of tactile feedback for precise manipulation. Our approach integrates both 3D vision and tactile feedback to enhance dexterous manipulation.

\subsection{Tactile for Dexterous Manipulation}
Tactile sensing has been widely studied for dexterous manipulation~\cite{qi2023general, guzey2023dexterity, wu2024canonical, yin2023rotating}. Vision-based tactile sensors~\cite{yuan2017connecting, lambeta2020digit, lambeta2024digitizing, padmanabha2020omnitact} offer high-resolution surface geometry but are typically large and hard to integrate. Piezoresistive sensors~\cite{li2019connecting, huang20243dvitac} are compact and robust, though limited to single-axis force detection. Magnetic sensors~\cite{tomo2017covering, bhirangi2024anyskin, bhirangi2021reskin}, which we adopt in our work, provide continuous tri-axial force feedback and are well-suited for integration into dexterous hands.
Recent learning-based tactile manipulation research has emphasized both pretraining and improved tactile representations to enhance learning efficiency and generalization. Pretraining~\cite{grill2020bootstrap} reduces reliance on task-specific datasets by learning transferable features from large-scale data such as play demonstrations~\cite{guzey2023dexterity}. Meanwhile, diverse tactile representations, such as 2D image representations~\cite{guzey2023dexterity, guzey2024see}, graph-based representations~\cite{funabashi2022multi, yang2023tacgnn}, and canonical representations~\cite{wu2024canonical}, have been proposed to capture structural relationships and improve downstream performance. However, most existing approaches rely on straightforward sensor feature combinations and do not fully explore more representative multimodal fusion for dexterous manipulation. In contrast, our work focuses on effectively fusing 3D tactile and 3D visual information to improve manipulation performance.

\subsection{Visual Tactile Fusion}
Recent studies increasingly leverage multisensory inputs to enhance robotic perception and manipulation~\cite{lee2019making, liu2024maniwav, yin2023rotating} Many existing approaches perform fusion at the raw data level by directly concatenating tactile signals with other modalities~\cite{yin2023rotating, yin2024learning}. 
While effective for low-dimensional inputs, this becomes challenging for high-dimensional data, which requires more careful design. For example, some methods~\cite{huang20243dvitac,yuan2024robot} use forward kinematics to transform taxel positions into the same coordinate frame as the point cloud, enabling raw-level fusion in a unified 3D space. Although this spatially aligns the data, it lacks integration at the feature level. 

Instead of fusing raw inputs directly, some methods~\cite{guzey2024see, wu2024canonical, guzey2023dexterity} encode each modality separately into a latent representation before fusion. These representations are either combined through simple operations such as concatenation~\cite{wu2024canonical}, or used for self-supervised pretraining to improve visuo-tactile integration. Pretraining strategies include bi-directional cross-modal prediction~\cite{li2019connecting}, contrastive learning on paired data~\cite{george2024visuo,dave2024multimodal,kerr2022self}, and modality reconstruction via transformer-based architectures~\cite{chenvisuo,liu2024masked}, which enhance cross-modal alignment and representation learning.
However, most of these methods focus on learning a comprehensive latent representation of the current observation, without considering that visual and tactile features vary in importance throughout different manipulation stages~\cite{badde2020modality}. In some cases, tactile feedback can even introduce noise or act as a distraction~\cite{he2024foar}. FoAR~\cite{he2024foar} addresses this by predicting contact probability to explicitly adjust the weight of tactile features during different manipulation stages, but it requires manual labeling and assumes visual features are always dominant. In contrast, we propose a general and efficient approach that avoids manual labeling, adaptively adjusts modality weights, and improves policy performance in dexterous manipulation tasks.

\section{ROBOT SYSTEM SETUP}
Our system integrates a 7-DoF Flexiv Rizon 4 robot arm and a customized 16-DoF Leap Hand~\cite{shaw2023leaphand} dexterous hand featuring four fingers. PaXini tactile sensors are distributed across each finger: one sensor on the fingertip and another on the fingerpad. Both sensor types feature a 3×5 array of taxels, with each taxel capable of measuring tri-axial forces $\force \in \mathbb{R}^3$. A single Intel RealSense L515 camera is mounted diagonally on the robot to capture visual information.

For expert demonstration collection, we utilize an additional Intel RealSense D415 camera with HaMeR~\cite{pavlakos2024reconstructing} to track human hand pose, enabling robot teleoperation through Dexpilot~\cite{handa2020dexpilot, qin2023anyteleop} retargeting. The robot arm is controlled via a target end-effector pose comprising 3D translation and 6D rotation representation~\cite{zhou2019continuity}, while the dexterous hand receives 16-DoF target joint position commands. Both demonstration collection and inference operate at a frequency of 5 Hz.

\section{METHOD}

\begin{figure*}[t!]
    \centering
    \vspace{2mm}
    \includegraphics[width=\linewidth]{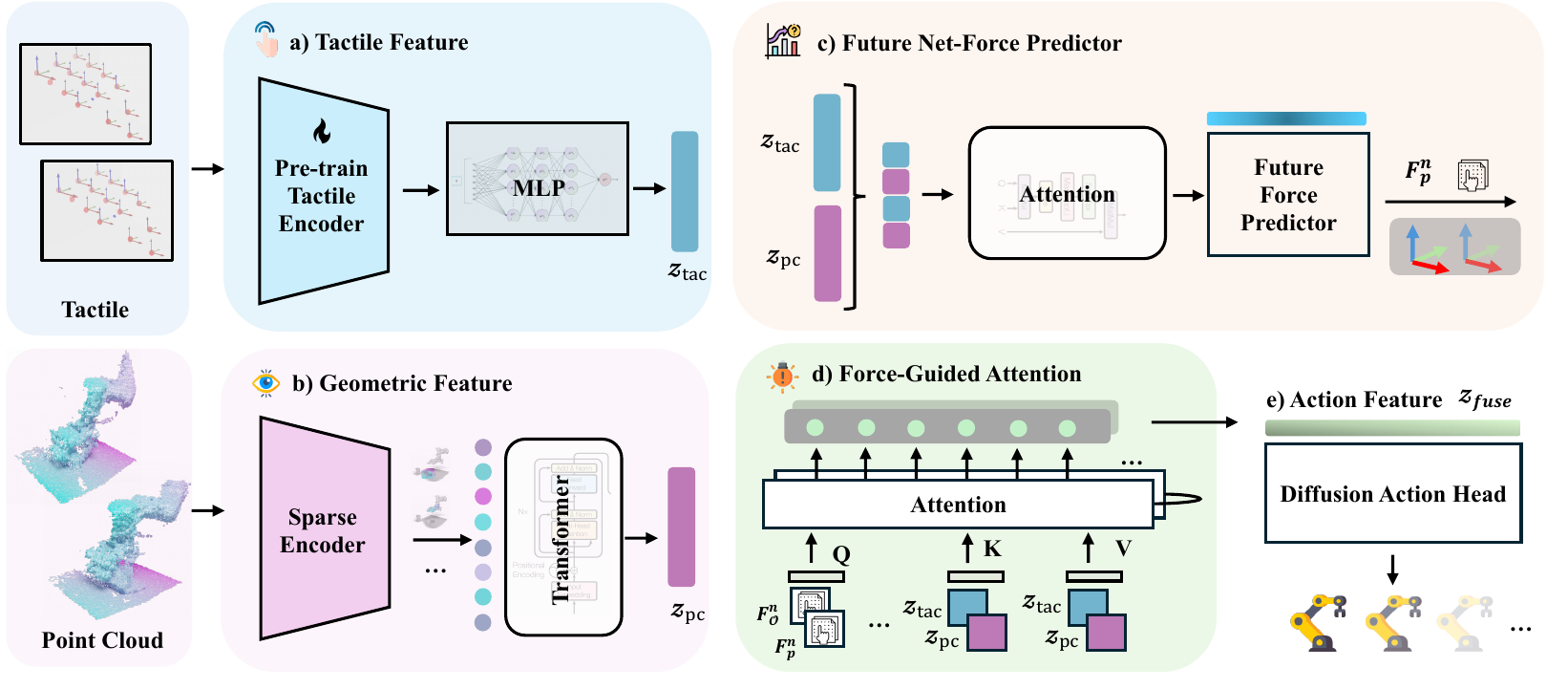}
    \caption{\textbf{Pipeline.} a) We use pretrained tactile encoder to encode 3D tactile. b) We use sparse encoder to encode the point cloud. c) The encoded visual and tactile features are used to predict the future net force. d) The predicted future net force is combined with the observed net force to guide visuo-tactile fusion through an attention mechanism. e) The fused action feature is used as a condition for learning the dexterous manipulation policy.}
    \label{fig:pipeline}
    \vspace{-5mm}
\end{figure*}

\textbf{Problem Statement}: We focus on the problem of fusing tactile data from distributed tactile 
sensors with point cloud data to learn adaptive visuo-tactile dexterous manipulation policies. 
Given a sequence of observations $\{\obs_{t-h+1}, \ldots, \obs_t\}$ over a history horizon $h$, 
where each observation $\obs_t = \{\obs_t^{\pc}, \obs_t^{\tact}\}$ consists of point cloud $\obs_t^{\pc} \in \mathbb{R}^{N \times 3}$ and tactile inputs $\obs_t^{\tact} \in \mathbb{R}^{120 \times 12}$~\cite{wu2024canonical},
our objective is to learn a policy $\pi$ that predicts the next $n$ robot actions $A_t = \{{A_{t+1}, A_{t+2}, \ldots, A_{t+n}}\}$. 
Each action $A_t \in \mathbb{R}^{25}$ consists of robot’s translation $\mathbf{t} \in \mathbb{R}^3$, rotation $\mathbf{r} \in \mathbb{R}^6$, and hand joint positions $\mathbf{q} \in \mathbb{R}^{16}$.

In this section, we first introduce our force-guided attention fusion module (Section~\ref{sec:atten}), and adaptively fuse visual and tactile features across different stages of manipulation. We then introduce a 
self-supervised future force prediction task, which combines predicted future force with observed force to guide attention modulation (Section~\ref{sec:fforce}). Finally, we integrate these components into an imitation learning framework for visuo-tactile policy training (Section~\ref{sec:policy-learning}). The overview of our architecture is shown in Fig.~\ref{fig:pipeline}.

\subsection{Force-Guided Attention Fusion}\label{sec:atten}
The force signal varies consistently with the stage of manipulation and thus serves as a natural signal for guiding attention.  
We use a sparse encoder~\cite{lee2019making} to extract point cloud features $\feat^\pc = \encoder_\pc(\obs^\pc)$ and a pretrained tactile encoder~\cite{wu2024canonical} to extract tactile features $\feat^\tact = \encoder_\tact(\obs^\tact)$.  
To incorporate force information, we then project the observed net force $\netforce_{\mathcal{O}}$ to $e^{\force} = \mlp_\force(\netforce_{\mathcal{O}})$ using an MLP $\mlp_\force$, where $\netforce_{\mathcal{O}}$ is the sum of tri-axial forces from all taxels, each transformed into the camera coordinate frame using its 6D pose. 

To align the feature dimensions, we apply separate MLPs to the point cloud and tactile features:  
$e^{\pc} = \mlp_{\pc}(\feat^{\pc}) \in \mathbb{R}^{512}, \, e^{\tact} = \mlp_\tact(\feat^{\tact}) \in \mathbb{R}^{512}$.
Based on these aligned features, we obtain the query, key, and value representations as follows:
\begin{equation}
\begin{aligned}
    \query^{\force} &= e^{\force} \weightmat_Q, \\
    \key &= [e^{\pc}, e^{\tact}] \weightmat_K, \\
    \val &= [e^{\pc}, e^{\tact}] \weightmat_V,
\end{aligned}
\end{equation}
where $\weightmat_Q$, $\weightmat_K$, and $\weightmat_V$ are learnable projection matrices, and $[\cdot]$ denotes feature concatenation. 

The attention mechanism first computes fusion weights between the force-guided query and the keys from visual and tactile features:
\begin{equation}
    \weight^{\pc}, \weight^{\tact} =
    \sigma\left(\frac{\query^{\force} \key^\mathrm{T}}{\sqrt{d_k}}\right),
\end{equation}
where $\weight$ denotes the attention weights, $\sigma(\cdot)$ is the softmax function, and $d_k$ is the dimensionality of the key vectors. The resulting weights are then applied to the corresponding values to produce the fused representation:
\begin{equation}
    \feat^{\text{fuse}} =
    \weight^{\pc} \val^{\pc} +
    \weight^{\tact} \val^{\tact},
\end{equation}
where $\feat^{\text{fuse}}$ denotes the fused feature.

This module enables adaptive weighting of visual and tactile features based on force, allowing the policy to prioritize touch when needed rather than always relying on vision~\cite{he2024foar}.

\subsection{Future Force Prediction and Guidance}\label{sec:fforce}
The proposed force-guided attention fusion enables adaptive weighting of modalities. However, without explicit supervision, it may not fully exploit each modality at the right stage of manipulation. For example, in a reorientation task, the robot primarily relies on visual input during the reaching phase, while both visual and tactile inputs are critical during the reorientation phase. Due to the richer visual information, the attention module may become biased toward vision, leading to insufficient utilization of tactile feedback.

To address this issue, we design a self-supervised future force prediction task during training, where we introduce a transformer-based \cite{vaswani2017attention} diffusion head \cite{ho2020denoising} for future net force prediction. 
The diffusion force head takes the visual and tactile features as input and predicts the future net force $\netforce_p$ at the next $n$ steps. 

We then concatenate the observed net force $\netforce_{\mathcal{O}}$ and predicted future forces $\netforce_p$ to form the guide force $\netforce_g$, and project it as the query:
\begin{equation}
\query^{\force} = g_F([\netforce_{\mathcal{O}}, \netforce_p]) \weightmat_Q.
\end{equation}
where $g_F$ is an MLP that projects the force vectors into the query space, and $\weightmat_Q$ is a learnable projection matrix.
This query guides the attention module using both current and future contact information.


\subsection{Visuo-Tactile Policy Learning}\label{sec:policy-learning}
We integrate the force-guided attention fusion and future force prediction into an imitation learning framework for visuo-tactile dexterous manipulation. 
Specifically, we adopt \textit{RISE}~\cite{10801678} as our base policy architecture, which is a 3D diffusion model-based policy. 
The extracted visual and tactile features are first used for future force prediction, 
and the predicted future force $\netforce_p$ is concatenated with the observed net force $\netforce_{\mathcal{O}}$ to form the guide force $\netforce_g$. 
This guide force is then used as the query in the cross-modal attention module. Finally, the fused features are passed to the diffusion action head~\cite{chi2023diffusionpolicy} 
to predict the future action. During the training, the policy loss $\mathcal{L}{\pi}$ combines with an additional future force prediction loss $\mathcal{L}_{\text{ffp}}$ to update the network.
The total loss is defined as:
\begin{equation} 
    \mathcal{L} = \mathcal{L}_{\pi} + \alpha\mathcal{L}_{\text{ffp}},
\end{equation} 
where $\alpha$ is a hyperparameter to adjust the future force prediction loss $\mathcal{L}_{\text{ffp}}$.

\section{EXPERIMENTS}

\begin{figure*}[h]
    \centering
    \vspace{2mm}
    \includegraphics[width=\linewidth]{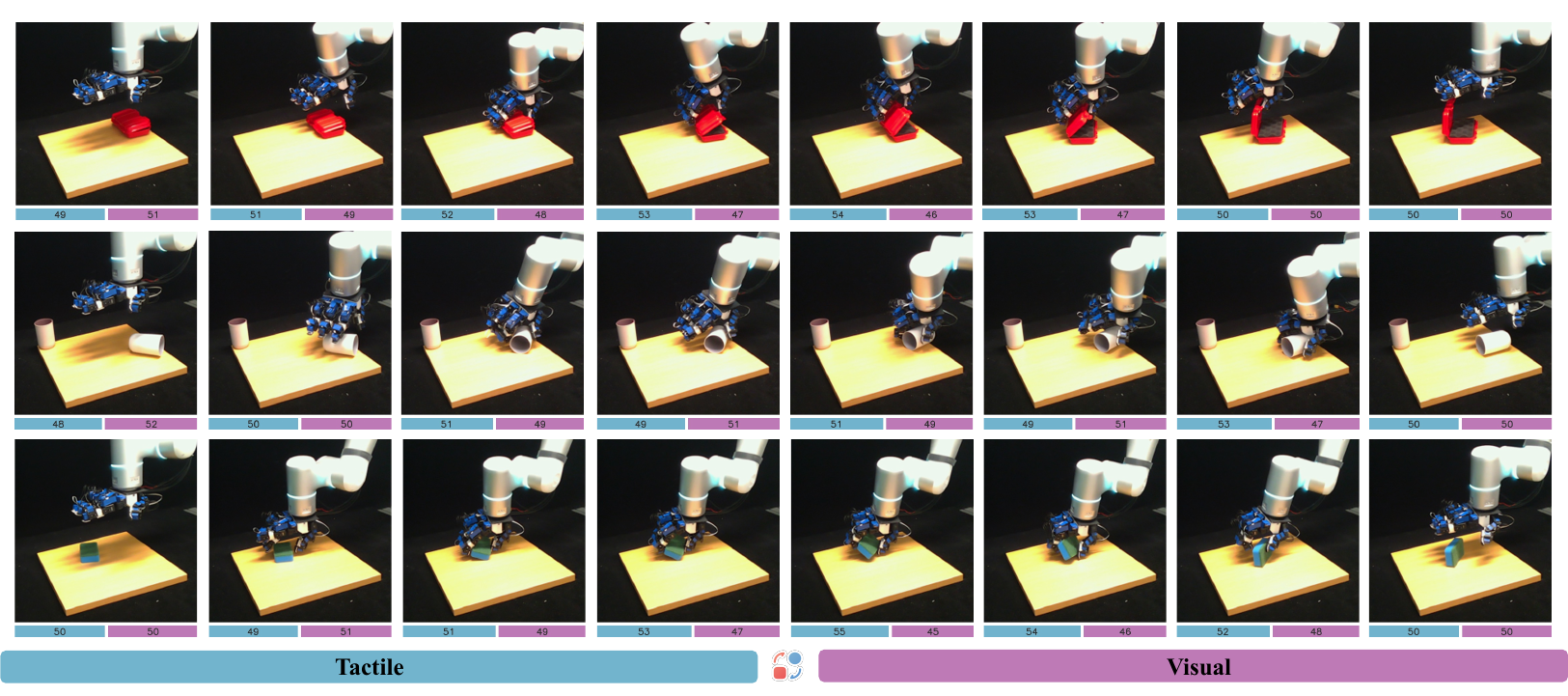}
    \vspace{-5mm}
    \caption{\textbf{Visualization of Our Policy's Rollout and Attention Weights on Three Contact-Rich Manipulation Tasks.} Note: this view corresponds to the robot's observation perspective, with point cloud data serving as the visual input. The bar below each image shows the attention weights assigned to the tactile (blue) and visual (purple) modalities, with the \textbf{numbers} indicating the exact weight values at each stage of manipulation.
    }
    \label{fig:tasks}
\end{figure*}

We conduct comprehensive real-world experiments to answer the following questions:
\begin{itemize} 
    \item Can our fusion module learn to adaptively use the visual and tactile features? 
    \item Does our future force prediction and guidance enhance the policy to learn appropriate attention adjustments? 
    \item How exactly does the attention focus during different manipulation stages of different tasks? 
 \end{itemize}

\subsection{Dexterous Manipulation Tasks}
We evaluate our approach on three dexterous, contact-rich manipulation tasks, as shown in \figmk\ref{fig:tasks}. For each task, we collect around 30 expert demonstrations to train the policy. During evaluation, we run 10 trials per method per task, with each trial limited to 300 steps. The initial object pose is randomized within a $35\times35$ cm planar workspace.

\textit{(i) Open Box}: The robot opens a box using the thumb, index, and middle fingers. It needs to first reach the box, grasp the upper part, and then adjust its fingers to open it. The key challenge is maintaining a firm hold on the upper part to prevent it from loosening and falling. Success is achieved if the upper part remains in place after opening.

\textit{(ii) Reorientation}: The robot reorients a cup to a target direction by coordinating four fingers. It needs to reach the cup and coordinate all four fingers to reorient it without pushing it out of the workspace, given the low surface friction. The challenge lies in precise finger coordination across a long horizon. Success is achieved if the final orientation is within $\pm10$ degrees of the target.

\textit{(iii) Flipping}: The robot flips a dish sponge using the thumb, index, and middle fingers. It needs to reach the sponge, lift one side, and flip it using the index finger. The challenge is precise finger coordination and force application under heavy visual occlusion. Success is defined as flipping the sponge upright by 90 degrees.

\subsection{Baselines}
We compare our method with the following three baselines. All methods share the same point cloud encoder, U-Net-based diffusion policy architecture, visual observations, and action space, differing only in their feature fusion approaches. 1) \textit{RISE}~\cite{10801678}: For this baseline, we implement its original policy, utilizing only point cloud data as input. 2) \textit{3DTacDex-P}~\cite{wu2024canonical}: We employ its pretrained encoder for tactile feature extraction, directly concatenating the visual and tactile features following the original implementation, and replacing its original RGB encoder with a sparse encoder~\cite{choy20194d} to ensure the same visual feature as ours. 3) \textit{FoAR}~\cite{he2024foar}: This baseline involves manually setting thresholds to label contact status and trains a predictor to estimate contact probabilities, which are used to weight tactile features. Following their approach, we determine suitable task-specific thresholds according to total force, use RGB and tactile data as inputs for contact prediction, and encode tactile data using the 3DTacDex~\cite{wu2024canonical} pretrained encoder.

\subsection{Manipulation Policy Comparison}
As shown in \tablemk\ref{table:main}, our proposed method (\textit{Ours}) outperforms all baselines across all tasks. 
The vision-only baseline, \textit{RISE}, performs well on tasks with strong visual information (Open Box, Reorientation) but struggles significantly on the Flip task. This task requires precise manipulation driven by tactile feedback, which \textit{RISE} lacks. This observation aligns with findings reported in 3DTacDex~\cite{wu2024canonical}. 
The \textit{3DTacDex-P} baseline, which employs visuo-tactile concatenation, demonstrates poor overall performance, even underperforming \textit{RISE}. We also observed that during training, \textit{3DTacDex-P} achieves lower hand-joint prediction error compared to other methods. These are likely due to overfitting to tactile patterns, which are similar across different expert demonstrations, while visuals varied a lot due to the larger workspace compared to the original method.
The \textit{FoAR} baseline, despite careful threshold selection for contact prediction, is unstable. While it performs well on the Reorientation task, it fails on both Open Box and Flip tasks. In these failures, the policy reaches the object but cannot apply the correct manipulation forces. Expert data reveals frequent contact changes, making manual thresholding unreliable and leading to inconsistent contact labels. 
In contrast, \textit{Ours} mitigates tactile overfitting by dynamically adjusting visuo-tactile attention across diverse tasks without relying on manual labeling, demonstrating robust generalization and consistent performance.

\begin{table}[h]
    \caption{\textbf{Success Rate of Different Manipulation Policies.}}
    \vspace{-5mm}
    \label{table:main}
    \begin{center}
        \resizebox{1\linewidth}{!}{
            \input{Tables/main.tex}

        }
    \end{center}
    \vspace{-5mm}
\end{table}

\subsection{Effectiveness of Force-Guided Attention Fusion}
To validate the effectiveness of force-guided attention fusion, we conduct experiments that do not include future force prediction and guidance. As shown in \tablemk\ref{table:method-ablation}, by incorporating only the attention module, the success rate increases from 40\% to 67\%. 
By observing the attention weights, we found that the policy learns to adjust the tactile and visual features during different manipulation stages instead of just overfitting, indicating the effectiveness of our proposed force-guided attention. However, although the policy learns to adjust the weights, we found that the visual feature is typically prioritized over the tactile feature across all tasks, likely due to the data imbalance. This observation explains the reason that the performance with the attention module alone is similar to \textit{RISE}, further highlighting the importance of future force prediction and guidance.


\subsection{Importance of Future Force Prediction and Guidance}
To demonstrate the importance of future force prediction (\textit{FFP}) and future force guidance (\textit{FFG}), we conduct ablation studies on the Flip task using different types of force prediction (\textit{FP-T}) and guided force (\textit{GF-T}). We also introduce a new metric, \textit{AEL}, representing the average episode length across all runs, with failed runs assigned the maximum episode length. 

As shown in \tablemk\ref{table:ffpg-ablation}, without both \textit{FFP} and \textit{FFG}, the success rate is only 50\%. Incorporating observed force prediction (\textit{OFP}) and guidance (\textit{OFG}) increases the success rate to 70\%, demonstrating the effectiveness of force prediction and guidance. Replacing observed force (\textit{OFP}) with future force prediction (\textit{FFP}) further raises the success rate to 90\%, similar to \textit{Ours}, highlighting the importance of predicting future force.
We also observe that for both \textit{OFP} and \textit{FFP}, more attention is given to the tactile modality upon contact, indicating that force prediction effectively reinforces tactile modality and improves balance. However, \textit{Ours w/o FFG} often requires multiple attempts, leading to an \textit{AEL} of 166, whereas \textit{Ours} typically succeeds on the first try. Additionally, \textit{Ours w/o FFG} exhibits more risky behaviors, such as continuously squeezing the board. These findings underscore the importance of \textit{FFG}.

\begin{table}[h]
    \caption{\textbf{Success Rate of Ablation.} \\ FGAF: our proposed force-guided attention fusion. FFPG: our proposed future force prediction and guidance. }
    \vspace{-5mm}
    \label{table:method-ablation}
    \begin{center}
        \resizebox{1\linewidth}{!}{
            \input{Tables/method_ablation.tex}
        }
    \end{center}
    \vspace{-7mm}
\end{table}


\begin{table}[h]  
    \caption{\textbf{Performance of Future Force Prediction and Guidance.} \\
     FP-T (Force Prediction Type); GF-T (Guided Force Type); SR (Success Rate); AEL (Average Episode Length); FFP (Future Force Prediction); FFG (Future Force Guidance); OFP (Observed Force Prediction); OFG (Observed Force Guidance).}
    \vspace{-5mm}
    \label{table:ffpg-ablation}
    \begin{center}
        \resizebox{1\linewidth}{!}{
            \input{Tables/ffpg.tex}
        }
    \end{center}
    \vspace{-7mm}
\end{table}

\subsection{Analysis of Attention}


We visualize the attention weights across different steps, as shown in \figmk\ref{fig:tasks}. 
For all tasks, during the reaching stages, more attention is given to the visual, as the hand manipulates the object, the attention shifts towards the tactile, indicating effective attention adjustment.
However, we observe that the changes in attention weights for the reorientation task are not substantial. Tactile features indeed gain more attention during contact, but visual attention remains relatively high. This is likely due to the importance of visual features not only during reaching but also throughout manipulation, as tasks like reorientation require vision to assess whether the object has been rotated to the correct angle. 

\subsection{Generalization on Unseen Objects}

\begin{figure}[t!]
    \centering
    \vspace{2mm}
    \includegraphics[width=\linewidth]{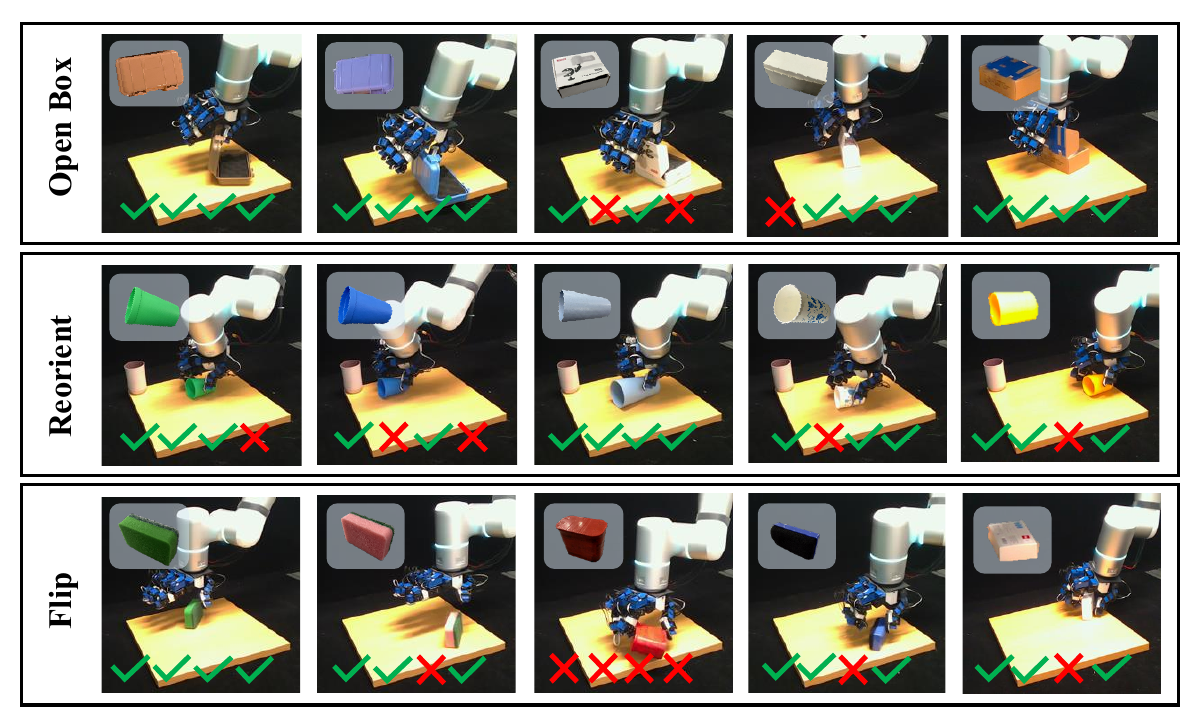}
    \caption{\textbf{Visualization of Our Policy on Unseen Objects.}}
    \label{fig:gen}
    \vspace{-5mm}
\end{figure}

We validate the generalization of our method by testing it on five objects with varying colors and geometries, as shown in \figmk\ref{fig:gen}. Each object is tested four times with different random poses. As shown in \tablemk\ref{table:method-generalization}, our policy achieves a 75\% success rate on unseen objects, demonstrating strong generalization, even for objects with significantly different geometries, such as a small paper cup in the reorientation task and a whiteboard eraser in the flipping task.

\begin{table}[h]
    \caption{\textbf{Success Rate of Unseen Objects.}}
    \vspace{-5mm}
    \label{table:method-generalization}
    \begin{center}
        \resizebox{1\linewidth}{!}{
            \input{Tables/genernalization.tex}

        }
    \end{center}
    \vspace{-5mm}
\end{table}

\section{CONCLUSIONS}

In this work, we enhance visuo-tactile fusion by proposing a force-guided attention fusion module that enables the policy to assign different levels of attention to different modalities at various manipulation stages in a more flexible manner, without human labeling. Additionally, we introduce a self-supervised future force prediction task and future force guidance to reinforce tactile modality, improving data imbalance and encouraging the policy to properly adjust attention weights, enhancing dexterous manipulation. Real-world experiments on three fine-grained, contact-rich dexterous tasks demonstrate the generalization of our method.

\textbf{Limitations and Future Work.}  
Although our method demonstrates strong generalization, it still cannot guarantee complete success on all tasks. Combining it with reinforcement learning could further improve robustness.

\section{Acknowledgments}
 We thank Hongjie Fang for insightful discussions, Tianshu Wu for assistance with camera calibration, and Mingjie Pan and Gu Zhang for discussions on point cloud networks. We also appreciate valuable feedback on the draft from Yujie Zhao, Hongwei Fan, Zhiyuan Ma, and Qiyang Yan, as well as the RISE/FoAR authors for their released code. This project was supported by the National Youth Talent Support Program (8200800081), National Natural Science Foundation of China (62376006) and National Natural Science Foundation of China (62136001).


\addtolength{\textheight}{-0cm}   
\bibliographystyle{IEEEtran}
\bibliography{ref}
\end{document}

%% file: notation.tex
\newcommand{\force}{\mathbf{F}}
\newcommand{\netforce}{\force^n}

\newcommand{\feat}{\mathcal{Z}}
\newcommand{\pc}{{pc}}
\newcommand{\tact}{{tac}}
\newcommand{\encoder}{\mathbf{\phi}}

\newcommand{\obs}{\mathcal{O}}

\newcommand{\weight}{\mathcal{\alpha}}
\newcommand{\weightmat}{W}

\newcommand{\query}{Q}
\newcommand{\key}{K}
\newcommand{\val}{V}

\newcommand{\mlp}{\mathbf{g}}

%% file: Tables/main.tex
\begin{tabular}{lccccc}

\toprule
\textbf{Method}                 & \textbf{Open Box} & \textbf{Reorientation} & \textbf{Flip}     & \textbf{Avg} \\ 
\midrule
\textbf{RISE}                   & 90\%              & 90\%                   & 40\%              & 73\%         \\
\textbf{3DTacDex-P}               & 30\%              & 60\%                   & 30\%              & 40\%         \\
\textbf{FoAR}                   & 20\%              &  90\%                   &  40\%              &  50\%        \\
\rowcolor{gray!20}\textbf{Ours} & \textbf{100\%}     & \textbf{90\%}          & \textbf{90\%}     & \textbf{93\%}  \\ 
\bottomrule
\end{tabular}

%% file: Tables/method_ablation.tex
\begin{tabular}{lccccc}
\toprule
\textbf{Method}                 & \textbf{Open Box} & \textbf{Reorientation} & \textbf{Flip}     & \textbf{Avg} \\ 
\midrule
\textbf{Ours w/o FFPG \& FGAF}   &  30\%             &  60\%                  &  30\%             &  40\%         \\
\textbf{Ours w/o FFPG}           &  80\%             &  70\%                  &  50\%             &  67\%        \\
\rowcolor{gray!20}\textbf{Ours} & \textbf{100\%}     & \textbf{90\%}          & \textbf{90\%}     & \textbf{93\%}  \\ 
\bottomrule
\end{tabular}

%% file: Tables/ffpg.tex


\begin{tabular}{lcccc}
\toprule
\textbf{Method}                              & \textbf{FP-T} & \textbf{GF-T} & \textbf{SR}     & \textbf{AEL} \\ 
\midrule
\textbf{Ours w/o FFP \& FFG}                 &  -                             & -                          &  50\%           &  221         \\
\textbf{Ours w/o FFP \& FFG, w OFP \& OFG}          &  obs                    & obs                        &  70\%           &  182         \\
\textbf{Ours w/o FFG, w OFG}                 &  future                        & obs                        &  90\%           &  166         \\
\rowcolor{gray!20}\textbf{Ours}              &  \textbf{future}               & \textbf{obs+future}        &  \textbf{90\%}  &  \textbf{113} \\ 
\bottomrule
\end{tabular}

%% file: Tables/genernalization.tex
\begin{tabular}{lccccc}
\toprule
\textbf{Method}      & \textbf{Open Box} & \textbf{Reorientation} & \textbf{Flip}  & \textbf{Avg} \\ 
\midrule
\textbf{Ours}     &  85\%                    &  75\%              &  65\%         &  75\%              \\
\bottomrule
\end{tabular}